\documentclass[10pt,twocolumn,letterpaper]{article}

\usepackage{cvpr}              

\usepackage{amsmath,amssymb,amsfonts}
\usepackage{graphicx}
\usepackage{xcolor}
\usepackage{booktabs}
\usepackage{tabularx}
\usepackage{multirow}
\usepackage{multicol}
\usepackage{algorithm}
\usepackage{algpseudocode}
\usepackage{placeins} 
\algrenewcommand\algorithmicrequire{\textbf{Require:}}
\algrenewcommand\algorithmicensure{\textbf{Ensure:}}

\newcommand{\ignore}[1]{}

\usepackage{microtype}
\definecolor{cvprblue}{rgb}{0.21,0.49,0.74}
\usepackage[accsupp]{axessibility}  
\usepackage{flafter} 
\usepackage{dblfloatfix} 
\usepackage[pagebackref,breaklinks,colorlinks,allcolors=cvprblue]{hyperref}

\setlist[itemize]{leftmargin=*,topsep=2pt,itemsep=0pt,parsep=0pt,partopsep=0pt}
\def\paperID{WAD-2}
\def\confName{CVPR}
\def\confYear{2026}

\title{Localization-Guided Foreground Augmentation in Autonomous Driving}

\author{
Jiawei Yong$^{1}$\thanks{Corresponding author.} \quad
Deyuan Qu$^{2}$ \quad
Qi Chen$^{2}$ \quad
Kentaro Oguchi$^{2}$ \quad
Shintaro Fukushima$^{1}$\\
$^{1}$Toyota Motor Corporation, Japan \quad $^{2}$Toyota Motor North America, USA\\
{\tt\small jiawei\_yong@mail.toyota.co.jp, s\_fukushima@mail.toyota.co.jp}\\
{\tt\small deyuan.qu@toyota.com, qi.chen@toyota.com, kentaro.oguchi@toyota.com}
}

\begin{document}
\maketitle

\begin{abstract}
%
Autonomous driving systems often degrade under adverse visibility conditions—such as rain, nighttime, or snow—where online scene geometry (e.g., lane dividers, road boundaries, and pedestrian crossings) becomes sparse or fragmented. While high-definition (HD) maps can provide missing structural context, they are costly to construct and maintain at scale.
We propose Localization-Guided Foreground Augmentation (LG-FA), a lightweight and plug-and-play
inference module that enhances foreground perception by enriching geometric context online. LG-FA: (i) incrementally constructs a sparse global vector layer from per-frame Bird’s-Eye View (BEV) predictions; (ii) estimates ego pose via class-constrained geometric alignment, jointly improving localization and completing missing local topology; and (iii) reprojects the augmented foreground into a unified global frame to improve per-frame predictions.
Experiments on challenging nuScenes sequences demonstrate that LG-FA improves the geometric completeness and temporal stability of BEV representations, reduces localization error, and produces globally consistent lane and topology reconstructions. The module can be seamlessly integrated into existing BEV-based perception systems without backbone modification. By providing a reliable geometric context prior, LG-FA enhances temporal consistency and supplies stable structural support for downstream modules such as tracking and decision-making.
\end{abstract}



\section{Introduction}
\begin{figure}[t]
    \centering
    \includegraphics[width=0.9\linewidth]{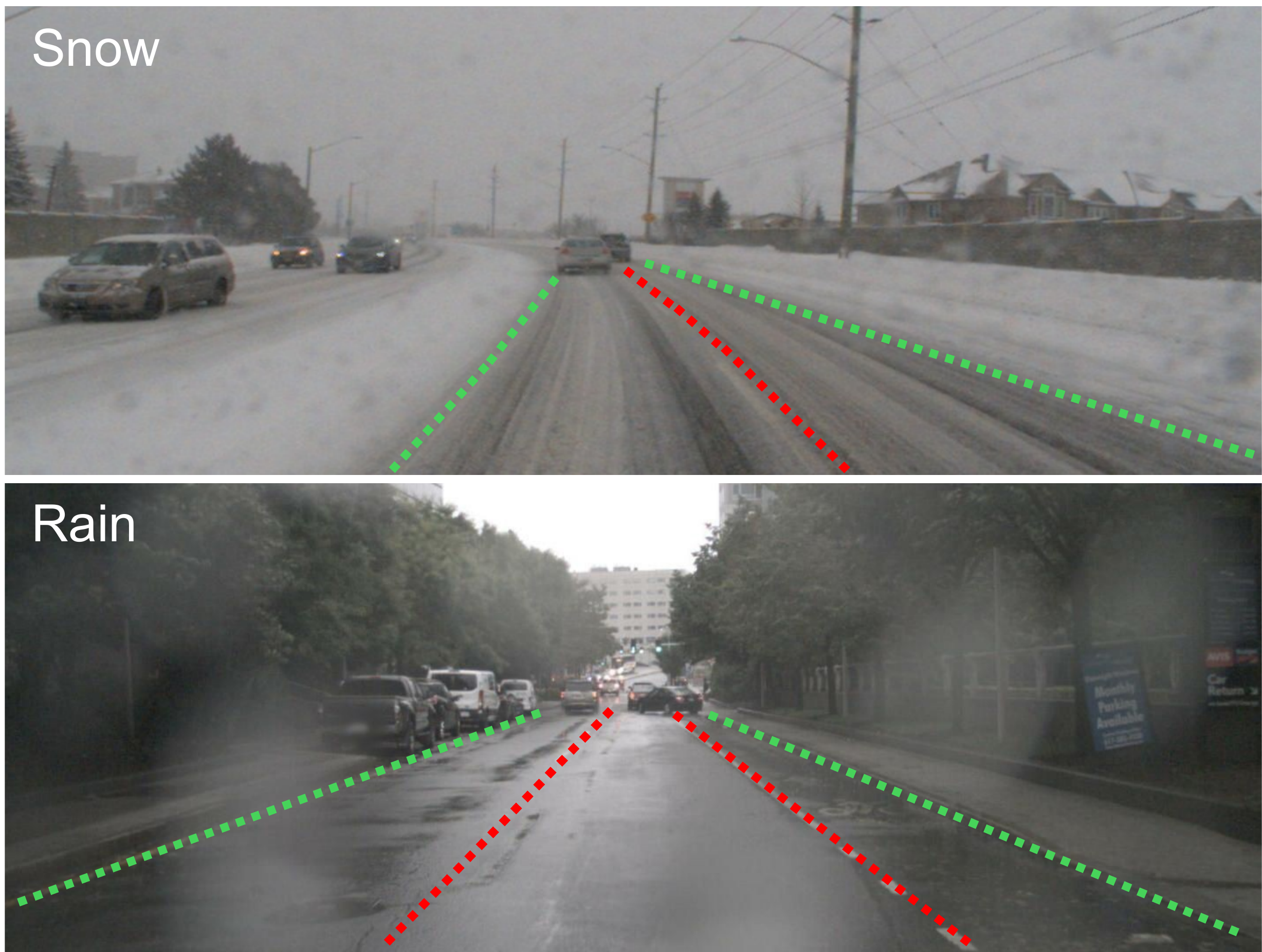}
    \caption{Challenging scenes under snow and rain where lane markings are faint or missing. Our localization-guided foreground augmentation (LG-FA) aligns the current frame with a lightweight global vector map to complete continuous lane dividers (\textcolor{red}{red}) and road boundaries (\textcolor{green}{green}), forming a stable augmented foreground perception for downstream tasks.}
    \label{fig:1}
    \vspace{-2mm}
\end{figure}
Autonomous driving (AD) systems rely heavily on Bird’s-Eye View (BEV) perception to support localization and downstream decision making~\cite{hu2023planning,Jiang2023VAD, zheng2024genad, hu2022st, yang2024visual}. However, under adverse-visibility conditions such as rain, snow, fog, or nighttime illumination, online scene geometry—including lane dividers, road boundaries, and pedestrian crossings—often becomes sparse or fragmented ~\cite{zhang2023perception, yoneda2019automated}. The loss of such geometric cues leaves the vehicle with insufficient lane and boundary structure for accurate localization and stable downstream decision making. This degradation pattern is illustrated in Fig.~\ref{fig:1}.

Although training-time robustness techniques, such as synthetic multi-weather data augmentation and cross-domain adaptation for perception, can alleviate perception degradation under adverse conditions~\cite{marathe2023wedge,li2023domain,zhao2024revisiting}, they mainly operate at the image or feature level to improve the robustness of perception models across domains. These methods enhance visual generalization for adverse-weather scenes, but they do not explicitly reconstruct persistent global road geometry or recover missing lane and boundary topology at runtime. Consequently, when structural cues such as lane dividers or boundaries become invisible, the BEV representation may still lack geometric completeness and spatial consistency.

A more effective strategy under such conditions is to enhance foreground perception by recovering a lightweight geometric scene context that restores road structure and improves temporal consistency. This can be achieved in two main ways: (i) High-definition (HD) maps, which provide centimeter-level geometry and topology but incur prohibitive construction and maintenance costs, making large-scale deployment impractical~\cite{elghazaly2023high, ebrahimi2022high, tang2023thma}; and (ii) Lightweight online maps constructed from the outputs of a BEV perception network, which provide a scalable and adaptive geometric prior. By localizing the current frame to such an adaptive geometric context and completing missing line structures online, the system can obtain a stable global reference for foreground perception without relying on expensive map infrastructure.
%


To enable this form of runtime foreground augmentation, we propose \textbf{Localization-Guided Foreground Augmentation (LG-FA)}, a lightweight module that constructs and leverages online geometric priors for foreground perception under degraded visibility. LG-FA first incrementally aggregates per-frame BEV geometry into a sparse \emph{global vector map}. It then performs class-constrained localization and line completion to align the current frame to this prior and recover missing local line structures, yielding a \emph{line-completed vector map} in the global frame. Finally, LG-FA reprojects the current-frame ego vehicle and detected objects into this line-completed vector map to form an \emph{Augmented Foreground Perception} representation with a shared global reference frame. LG-FA requires no backbone modification, can be attached at inference time. 
By converting completed geometric context into an enhanced foreground perception representation, LG-FA provides a practical and scalable way to improve perception reliability in degraded environments and offers a more stable input for downstream tasks.

The contributions of this work are as follows:
\begin{itemize}

\item We propose LG-FA, a lightweight plug-and-play runtime module that enhances foreground perception under degraded visibility using map geometric priors, without modifying the backbone network.

\item We introduce a pipeline with global vector map construction, class-constrained localization, line completion, and foreground reprojection onto a line-completed vector map to form an augmented foreground perception.

\item Experiments on nuScenes dataset show that LG-FA enables high-quality global map construction and accurate localization and line completion under degraded visibility, thereby improving the geometric reliability and stability of foreground perception.
\end{itemize}

\begin{figure*}[t]
    \centering
     \includegraphics[width=\textwidth]{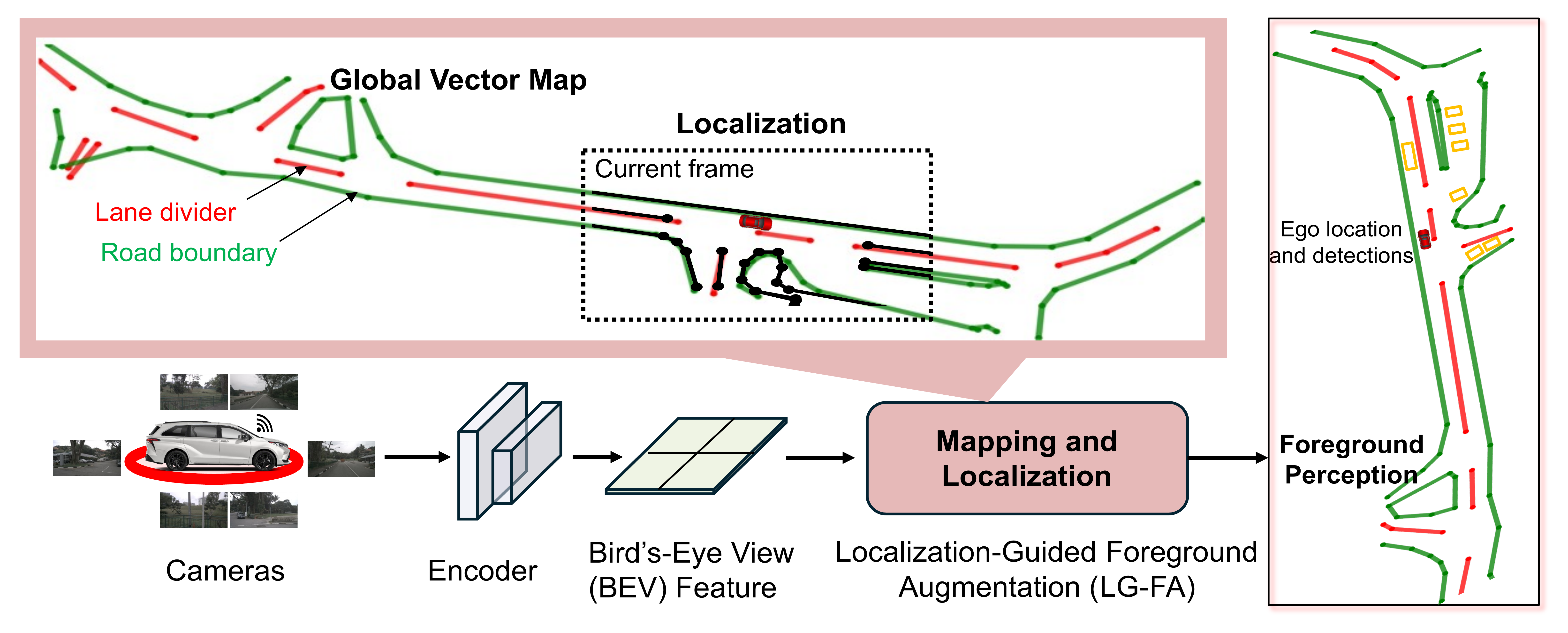}
    \caption{Overall architecture of the proposed Localization-Guided Foreground Augmentation (LG-FA) framework. Multi-camera inputs are first encoded into a Bird’s-Eye View (BEV) representation. LG-FA incrementally constructs a sparse global vector map from per-frame lane and topology predictions in an incremental aggregation stage. During online inference, the constructed map is used to estimate ego pose via class-constrained geometric alignment, jointly improving lane-level localization accuracy and completing missing local topology. The augmented foreground representations are then fed to downstream modules for motion forecasting and decision-making. This localization-guided augmentation bridges online perception with lightweight map priors, helping maintain temporal consistency when per-frame geometric cues are sparse or fragmented.}
    \label{fig:2}
    \vspace{-1mm}
\end{figure*}

\section{Related Work}
\label{sec:related}

\subsection{Bird's-Eye View (BEV) Perception}

Recent BEV perception has become a core representation for autonomous driving and is widely used for unified scene understanding across perception, forecasting, and planning~\cite{hu2023planning,Jiang2023VAD,zheng2024genad,hu2022st,yang2024visual}. Representative BEV-based systems include UniAD~\cite{hu2023planning}, VAD~\cite{Jiang2023VAD}, ST-P3~\cite{hu2022st}, ViDAR~\cite{yang2024visual}.
Under adverse conditions (e.g., rain, snow, fog, and nighttime), many BEV systems rely on training-time robustness techniques such as data augmentation and domain adaptation~\cite{marathe2023wedge,li2023domain,zhao2024revisiting}. While these methods improve robustness to appearance changes, they mainly depend on cues from current observations or short temporal windows. When lane markings and road boundaries become weak or invisible, BEV models typically lack an explicit geometric prior to recover missing structures at test time, motivating map-like geometric context as a more stable source of structural cues.

\subsection{High-definition Map and Online Map Methods}

A common solution for missing online geometry is to use high-definition (HD) maps as a structural prior, together with map-based localization to align current observations to the map frame. Classical methods include NDT~\cite{biber2003normal}, ICP~\cite{besl1992method}, and visual SLAM~\cite{campos2021orb}; cooperative LiDAR-based localization can further refine relative poses~\cite{dong2023lidar}. These methods can achieve high accuracy, but HD map construction, maintenance, and updates are costly~\cite{Lyu2025Survey,elghazaly2023high,ebrahimi2022high,tang2023thma}.

Recent works therefore study online vectorized map construction from onboard sensors, including HDMapNet~\cite{Li2022HDMapNet}, VectorMapNet~\cite{Liu2023VectorMapNet}, StreamMapNet~\cite{Yuan2024StreamMapNet}, MapTR~\cite{liao2022maptr}, MGMap~\cite{Liu2024MGMap}, and MapUnveiler~\cite{Kim2024MapUnveiler}. These methods show strong online mapping performance and the value of explicit geometric structure. However, their primary focus is online map reconstruction itself, rather than runtime foreground perception enhancement under degraded visibility. In contrast, our work reuses BEV geometric outputs from existing perception networks to build a lightweight global geometric prior and enhance foreground perception through localization-guided line completion.




\section{Methodology}
This section presents the \textbf{Localization-Guided Foreground Augmentation (LG-FA)} module, a lightweight plug-and-play runtime component that improves the geometric continuity, temporal stability, and spatial consistency of foreground perception under degraded visibility conditions. Rather than modifying the backbone architecture of an existing BEV-based perception network, LG-FA constructs and leverages a line-completed vector map context to produce more reliable foreground perception. As shown in Fig.~\ref{fig:2}, the module consists of three major components: (1)~\textit{Global Vector Map Construction}, which incrementally aggregates per-frame BEV geometry into a sparse global vector map that serves as a lightweight geometric prior; (2)~\textit{Localization and Line Completion}, which aligns the current frame to this prior and restores missing local line geometry in the global frame while preserving topology; and (3)~\textit{Augmented Foreground Perception}, which reprojects the current-frame ego vehicle and detected objects into the line-completed vector map to form an augmented foreground perception representation with a shared global reference frame. This representation can provide a more stable and geometrically consistent input for downstream tasks.

\subsection{Global Vector Map Construction}
We attach an online global mapping module on top of an existing BEV-based perception network. At each time step, the network produces a frame-level vectorized map, and our module incrementally fuses these vectors into a sparse scene-level global vector map that serves as a lightweight geometric prior for subsequent localization and line completion.

\textbf{Problem formulation.}
At time $t$, for each class $c\in\{\text{pedestrian crossing},\,\text{lane divider},\,\text{road boundary}\}$,
the network outputs a set of predicted polylines, optionally with persistent IDs:
\begin{subequations}\label{eq:vt_polyline}
\begin{align}
\mathcal{V}_t^c &= \{(\Gamma_{t,i}^c,\ \mathrm{id}_{t,i}^c)\}_{i=1}^{N_t^c}, \label{eq:vt_set}\\
\Gamma_{t,i}^c &= [p_1,\dots,p_{K_i}],\quad p_k\in\mathbb{R}^2. \label{eq:vt_poly}
\end{align}
\end{subequations}
Here, $\mathcal{V}_t^c$ denotes the per-frame vectorized map output of class $c$ at time $t$, and each element
$(\Gamma_{t,i}^c,\mathrm{id}_{t,i}^c)$ consists of the polyline geometry and an optional persistent ID.
$\Gamma_{t,i}^c$ is the $i$-th predicted polyline of class $c$, represented as an ordered sequence of 2D vertices in the ego-centric BEV frame, with the vertex order following the polyline arc length. $N_t^c$ is the number of predicted polylines of class $c$ at time $t$, and $K_i$ is the number of vertices in the $i$-th polyline.

We use an externally provided ego pose reference $\hat T_t$ to transform predicted polylines from the ego-centric BEV frame to a unified global scene frame. Specifically, each polyline is transformed as
\begin{equation}
\label{eq:transform}
\tilde\Gamma_{t,i}^c = \hat T_t \circ \Gamma_{t,i}^c .
\end{equation}
For each class, we maintain a global set of scene-level polylines
$\mathcal{G}^c=\{G_j^c\}_{j=1}^{M^c}$ and an online association
$a_t^c:\{1,\dots,N_t^c\}\rightarrow \{1,\dots,M^c\}\cup\{\varnothing\}$.
Here, $M^c$ is the current number of global polylines for class $c$.
$a_t^c(i)=j$ indicates that the current polyline $\tilde\Gamma_{t,i}^c$ is associated with the $j$-th global element $G_j^c$, and $a_t^c(i)=\varnothing$ indicates no match.

To measure geometric consistency between a current-frame polyline and a global polyline, we define a symmetric nearest-neighbor discrepancy on sampled points.
Let $S_\delta(\cdot)$ denote arc-length sampling with step size $\delta$.
The directed average distance is
\begin{equation}
\label{eq:directed_distance}
\overline d(\tilde\Gamma\!\to\!G) =
\frac{1}{|S_\delta(\tilde\Gamma)|}
\sum_{x\in S_\delta(\tilde\Gamma)}
\min_{y\in S_\delta(G)} \|x-y\| ,
\end{equation}
and the symmetric discrepancy is
\begin{equation}
\label{eq:symmetric_distance}
D(\tilde\Gamma, G) =
\tfrac12\!\left(
\overline d(\tilde\Gamma\!\to\!G)
+ \overline d(G\!\to\!\tilde\Gamma)
\right) .
\end{equation}
Here $\tilde\Gamma$ and $G$ denote a transformed current-frame polyline and a candidate global polyline, respectively. This discrepancy is used in the subsequent same-class association and geometry-preserving map update.

\textbf{Online mapping.}
For each class $c$, we incrementally update the global vector map using the current-frame polylines in $\mathcal{V}_t^c$.
(1) \emph{Coordinate unification:} Each current-frame polyline $\Gamma_{t,i}^c$ is transformed to the unified global scene frame using the ego pose reference, yielding $\tilde\Gamma_{t,i}^c$.
(2) \emph{Association:} If a persistent ID is available, the current polyline is associated with the corresponding global element by ID. Otherwise, same-class geometric association is performed by matching the current polyline to the best-matching global polyline using the discrepancy defined above, subject to an association threshold. If no candidate satisfies the threshold, a new global element is initialized. Very short one-frame fragments are suppressed to reduce spurious map growth.
(3) \emph{Incremental merge:} For each matched pair, the associated global element is updated by incorporating samples from the current polyline and removing near-duplicate samples within a small tolerance, optionally with local averaging. This preserves local geometric consistency while controlling map size.
(4) \emph{Topology preservation:} To bridge small gaps, nearby endpoints are snapped only under class and locality constraints, and only when endpoint distance and tangent-direction consistency satisfy predefined thresholds. This reduces fragmentation while avoiding erroneous over-connection.
(5) \emph{Resampling and simplification:} Near-uniform arc-length resampling is followed by polyline simplification to obtain compact yet geometrically faithful global curves.
Finally, the union $\mathcal{G}=\bigcup_c \mathcal{G}^c$ forms the sparse global vector map used for subsequent localization and line completion.

\subsection{Localization and Line Completion}
Given the frame-level vectors $\mathcal{V}_t^c=\{(\Gamma_{t,i}^c,\mathrm{id}_{t,i}^c)\}$ and the global polylines $\mathcal{G}^c=\{G_j^c\}$ for classes
$c\in\{\text{pedestrian crossing},\,\text{lane divider},\,\text{road boundary}\}$,
we process these 2D vectorized representations in two steps:
(i) class-constrained 2D localization to estimate the current pose alignment to the global vector map, and
(ii) line completion to recover missing local line geometry using the aligned global vector map prior.
In this section, $\mathcal{V}_t^c$ denotes the current frame to be localized and completed with respect to the previously constructed global vector map, and is not restricted to the frames used during the map construction.

\textbf{Class-constrained localization.}
We first resample all polylines with arc-length step $\delta$.
For each class $c$, this yields (i) a current-frame sampled point set $\mathbf{X}^c=\{x_n\}$ and
(ii) a sampled point set from the global vector map $\mathbf{Y}^c=\{y_m\}$.
In addition, for the global vector map we form a class-wise segment set $\mathcal{S}^c$ by converting each global polyline into line segments between adjacent sampled vertices.

Let $\theta=(t_x,t_y,\phi)$ denote the 2D pose alignment parameters, with planar rotation $R(\phi)$ and translation
$t=[t_x,t_y]^\top$.
Here, $\theta$ is the pose-alignment variable optimized in the localization stage, and the transformation maps the sampled points from the current-frame vector polylines to the global scene frame for correspondence search against the global vector map.
The transformed current-frame points are
\[
\hat{\mathbf{X}}^c(\theta)=\{R(\phi)x+t \mid x\in\mathbf{X}^c\}.
\]

To improve alignment robustness and reduce one-sided matching bias, we use a bidirectional objective with two complementary terms: a \emph{forward} term that aligns transformed current-frame observations to the global vector map, and a \emph{backward} term that enforces reverse consistency from the global map to the transformed current-frame observations.

For the \emph{forward} term, we measure point-to-segment discrepancy from transformed current-frame sampled points to line segments of the global vector map:
\begin{equation}
\label{eq:forward_p2seg_loss}
d_{\mathrm{f}}(\mathbf{X}^c,\mathcal{S}^c;\theta)
=
\frac{1}{|\mathbf{X}^c|}
\sum_{x\in\mathbf{X}^c}
\rho\!\left(
\min_{s\in\mathcal{S}^c}
d_{\mathrm{seg}}\!\big(R(\phi)x+t,\ s\big)
\right),
\end{equation}
where $d_{\mathrm{seg}}(p,s)$ denotes the Euclidean distance from point $p$ to line segment $s$, and $\rho$ is a robust loss.

For the \emph{backward} term, we use sampled points from the global vector map and measure point-to-point discrepancy from these sampled points to the transformed current-frame sampled points:

\begin{equation}
\label{eq:backward_p2p_loss}
d_{\mathrm{b}}(\mathbf{Y}^c,\mathbf{X}^c;\theta)
=
\frac{1}{|\mathbf{Y}^c|}
\sum_{y\in\mathbf{Y}^c}
\rho\!\left(
\min_{x\in\mathbf{X}^c}
\|R(\phi)x+t-y\|
\right).
\end{equation}

We estimate the pose by minimizing the class-weighted bidirectional objective:
\begin{equation}
\label{eq:bidirectional_objective_p2seg}
\hat{\theta}
=
\arg\min_{\theta}
\sum_c w_c\!\left[
d_{\mathrm{f}}(\mathbf{X}^c,\mathcal{S}^c;\theta)
+
d_{\mathrm{b}}(\mathbf{Y}^c,\mathbf{X}^c;\theta)
\right],
\end{equation}
where $w_c$ denotes the class weight.
Classes with empty or insufficient sampled points and segments are skipped.

In the Iterative Closest Point (ICP)-style solver, correspondence search is restricted within the same class and gated by a class-wise threshold $r_c$. Here, $w_c$ balances the contribution of different semantic classes in the objective, $r_c$ limits class-wise correspondence search in the solver, and $\rho$ suppresses the influence of outlier correspondences.
Eq.~\eqref{eq:bidirectional_objective_p2seg} is optimized approximately using the class-constrained alternating procedure in Algorithm~\ref{alg:loc_bev_ieee}. In Algorithm~\ref{alg:loc_bev_ieee}, \emph{BuildCorr} constructs class-wise gated correspondences using forward point-to-segment projection and backward point-to-point nearest-neighbor matching. \emph{RobustWeights} computes correspondence weights from residuals using class weights $\{w_c\}$ and robust loss $\rho$. \emph{WeightedProcrustes} denotes a weighted 2D rigid alignment update that estimates the rotation and translation minimizing the weighted squared residuals over the current correspondence set. \emph{PoseDiff} measures the pose update using a scaled norm that balances translation and rotation units. For brevity, $\{r_c\}$, $\{w_c\}$, and $\rho$ are omitted from the input list of Algorithm~\ref{alg:loc_bev_ieee}, but are used internally by \emph{BuildCorr} and \emph{RobustWeights}.

In our implementation, we further use a coarse-to-fine schedule: a coarse stage using only \emph{road boundary} geometry with a relaxed gate, followed by an all-class refinement stage with tighter class-wise gates. The final pose is obtained by composing the two-stage transforms.
\begin{algorithm}[t]
\caption{Class-Constrained Localization}
\label{alg:loc_bev_ieee}
\begin{algorithmic}[1]
\Require class-wise current sampled points $\{\mathbf{X}^c\}$, class-wise global sampled points $\{\mathbf{Y}^c\}$, class-wise global segments $\{\mathcal{S}^c\}$, initial pose $\theta_0$, max iterations $K$, convergence tolerance $\varepsilon_{\text{conv}}$
\Ensure estimated alignment pose $\hat{\theta}$

\State $\theta \gets \theta_0$
\For{$k=1$ to $K$}
  \State $\mathcal{P} \gets \mathrm{BuildCorr}(\theta)$
  \If{$\mathcal{P}=\varnothing$}
    \State \textbf{break}
  \EndIf

  \State $\mathbf{w} \gets \mathrm{RobustWeights}(\mathcal{P}, \theta)$
  \State $\theta_{\text{new}} \gets \mathrm{WeightedProcrustes}(\mathcal{P}, \mathbf{w})$

  \If{$\mathrm{PoseDiff}(\theta_{\text{new}}, \theta) < \varepsilon_{\text{conv}}$}
    \State $\theta \gets \theta_{\text{new}}$
    \State \textbf{break}
  \EndIf

  \State $\theta \gets \theta_{\text{new}}$
\EndFor
\State \Return $\hat{\theta} \gets \theta$
\end{algorithmic}
\end{algorithm}

\textbf{Line completion.}
After localization, we apply $\hat{\theta}$ to $\mathcal{V}_t^c$ and obtain transformed current-frame vectors in the global scene frame.
For each global polyline $G_j^c$, we evaluate coverage along its arc-length parameterization using segments induced by the transformed current-frame polylines within a narrow buffer around $G_j^c$, and identify uncovered intervals.

We then complete missing line geometry using a gap-length-dependent strategy.
Gaps shorter than a threshold $\eta$ are bridged directly, while longer intervals are completed by splicing the corresponding segments from the global polyline under endpoint continuity and tangent-direction continuity constraints.
This yields a line-completed vector map $\widetilde{\mathcal{V}}_t^c$ in the global scene frame, providing a more continuous and geometrically complete vector map for foreground perception.

\subsection{Augmented Foreground Perception}
\begin{figure}[t]
    \centering
    \includegraphics[width=0.9\linewidth]{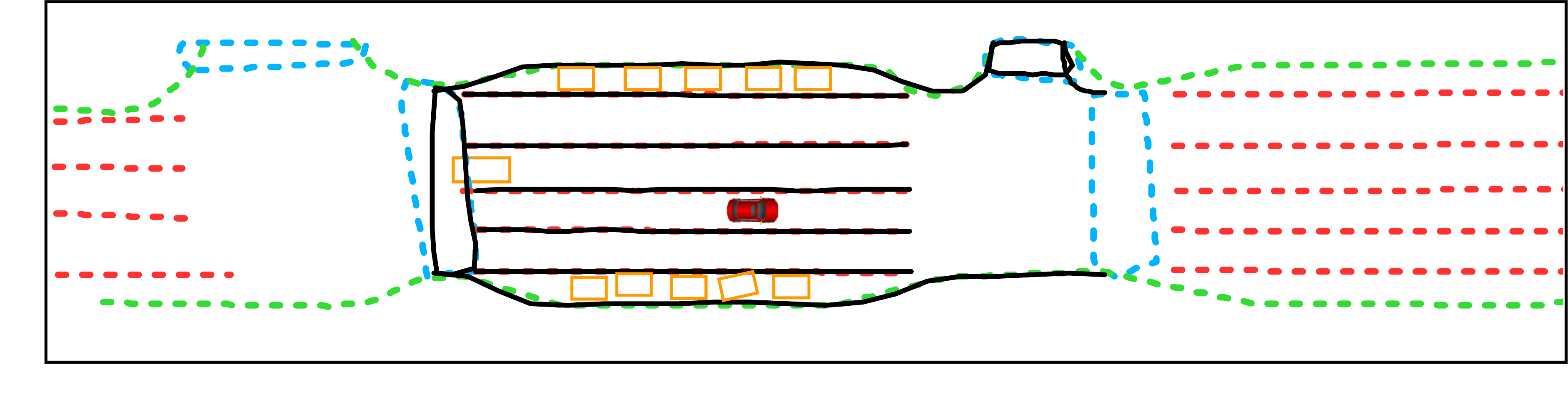}
    \vspace{-4mm}
    \caption{Ego position (\textcolor{red}{red icon}) and detected objects (\textcolor{orange}{yellow boxes}) reprojected onto the completed map.
Colored dashed curves denote the constructed global vector map, while black solid lines indicate the incomplete map predictions from the current frame.
Together they form the augmented foreground perception.}
    \label{fig:aug_foreground_perception}
    \vspace{-1mm}
\end{figure}

Once the line-completed vector map is obtained, LG-FA uses the refined ego pose estimated in the localization stage to reproject the current-frame ego vehicle and all detected objects into the coordinate frame of the completed map. In this way, the current perception results and the line-completed vector map share a common global reference frame. The ego position and object bounding boxes are then overlaid onto the completed lane dividers, road boundaries, and pedestrian crossings. We refer to this combined description as an \emph{augmented foreground perception}. Figure~\ref{fig:aug_foreground_perception} shows an example of an augmented foreground perception view, where the ego vehicle and detected objects are overlaid on the line-completed vector map.
The quality of the augmented foreground perception relies on both refined localization and map completeness. Refined localization improves geometric alignment accuracy, ensuring that the ego vehicle and detected objects are projected to more accurate and geometrically consistent locations in the map frame. The line-completed vector map, in turn, provides continuous lane-level and boundary-level context even when local observations are sparse or fragmented under degraded visibility conditions. Together, they yield a more accurate and stable geometric interpretation of foreground perception.

This augmented foreground perception provides a more stable structural interface for downstream processing. Under poor visibility, frame-wise perception in the original vehicle-centric frame is often sparse or noisy, leading to inconsistent scene interpretation across frames. After reprojection, both the ego state and surrounding objects are anchored to a continuous and globally consistent road geometry, so that subsequent modules can operate on a more robust scene frame instead of a drifting local frame with missing structure. In this work, we focus on improving current-frame foreground perception quality through geometric alignment and context completion, rather than explicitly modeling or evaluating downstream prediction or planning. 

\section{Experiments}
We conduct extensive experiments on LG-FA using the nuScenes dataset~\cite{caesar2020nuscenes}.
All experiments are conducted on a server equipped with an Intel Xeon Platinum 8468 CPU and 4$\times$ NVIDIA A100 80GB GPUs (CUDA 12.8).
%
First, we construct a global vector map by fusion of per-frame BEV vector predictions and compare it with ground truth to assess geometric consistency and scale stability. This fused map serves as a lightweight geometric prior for localization and line completion.
%
Second, we evaluate alignment and line completion under different initialization noise levels.
%
Third, we quantitatively evaluate foreground perception enhancement using a completion rate metric.
The complete source code can be made publicly available upon request or if required, while the backbone checkpoint may be subject to licensing constraints.
\begin{figure*}[t]
  \centering
\includegraphics[width=\textwidth,height=0.22\textheight,keepaspectratio]{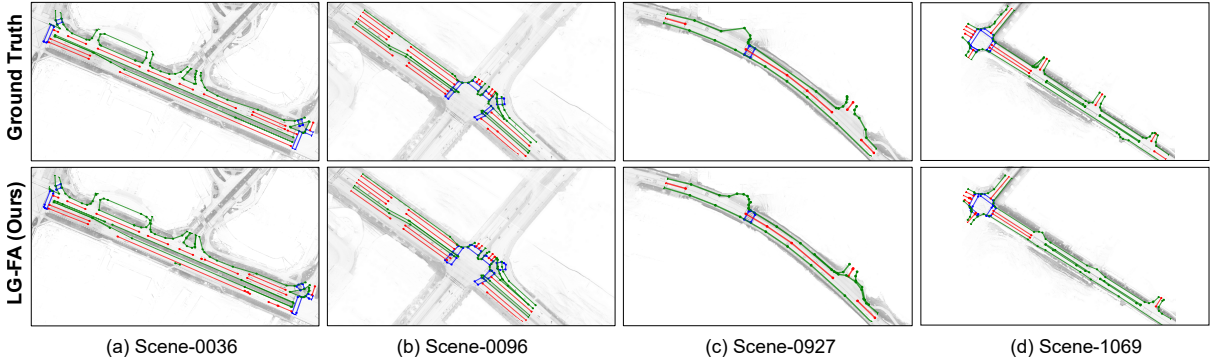}
  \caption{Qualitative comparison of our constructed global vector maps with ground truth across four scenes from nuScenes.}
  \label{fig:vis_map}
  \vspace{0mm}
\end{figure*}
\begin{table}[t]
\caption{Global vector map quality on the nuScenes validation split.
We report mean/median/90th percentile per class (computed on
scenes where both ground truth and prediction contain that class).}
\label{tab:map_table}
\setlength{\tabcolsep}{3.5pt}
\renewcommand{\arraystretch}{1.05}
\small
\resizebox{\columnwidth}{!}{%
\begin{tabular}{l c c c}
\toprule
Class & $N$ & Chamfer (m) & Scale Error (\%) \\
\midrule
Ped. crossing & 99 & 2.16/0.55/3.35 & 3.2/1.1/5.8 \\
Divider & 150 & 0.68/0.49/1.28 & 1.5/0.9/2.7 \\
Boundary & 149 & 0.68/0.53/1.22 & 1.7/1.0/3.1 \\
\bottomrule
\end{tabular}%
}
\vspace{0.5mm}
\end{table}
\subsection{Experiment Setup}
\textbf{Dataset.}  
We use the nuScenes dataset~\cite{caesar2020nuscenes}, strictly following the official \textit{train/val/test} split for training and evaluation. The mapping experiments cover both North American and Singapore city regions included in nuScenes. For localization robustness evaluation (pose refinement within mapping), since nuScenes provides accurate vehicle poses, we follow~\cite{dong2023lidar} and inject Gaussian noise with standard deviations $\sigma_{xy}=1\,\text{m}$ and $\sigma_\theta=2^\circ$, scaled by $\alpha\in[1,3]$ to control difficulty. In addition, we include representative rain and nighttime scenes from the official validation split for qualitative analysis under low-visibility conditions.

\textbf{Mapping baseline and metrics.}  
For mapping, we compare our constructed global vector map with a reference global map built from the official nuScenes vector map layers under the same alignment protocol, evaluating geometric consistency and scale stability.
We evaluate geometric consistency using \textit{Chamfer Distance}, following common practice in vectorized map evaluation exemplified by VectorMapNet~\cite{Liu2023VectorMapNet}.
\textit{Chamfer Distance} measures the average bidirectional point-wise deviation between the predicted and ground-truth polylines, and is representative of the overall geometric accuracy for the majority of scenes.
We further report the global \textit{Scale Error} in percent, defined as $\lvert 1 - s\rvert \times 100$, following ORB-SLAM3~\cite{campos2021orb}, where $s$ is the estimated scale factor after similarity alignment between the predicted and ground truth maps.


\textbf{Localization baseline and metrics.}
For localization, we use GNSS as the initialization baseline, and compare against ICP~\cite{besl1992method} and NDT~\cite{biber2003normal} as registration-based refinement baselines. All methods are evaluated under the same initialization noise levels and on the same data to ensure a fair comparison. We report translation error (m) and heading error (deg) as the primary accuracy metrics. In Table~\ref{tab:loc_table}, we evaluate frame-wise 2D rigid alignment on the ground plane (x, y, yaw) between the current-frame local vector map and the constructed global vector map.
The global vector map is loaded from the merged polyline JSON (represented in the last-frame ego coordinates) and is transformed into each target frame using nuScenes ego poses; thus the ground truth transform in the target-ego frame is identity.
The local vector map comes from the per-frame polyline prediction of the same scene and frame.
We densify polylines with a 0.2\,m step and cap the number of points for efficiency (global/local: 40k/20k; NDT source: 8k; frames with fewer than 30 points are skipped).
All methods start from the same noisy initialization (GNSS): for each noise scale $\alpha$, we apply $K\!=\!8$ deterministic perturbations (translations in $\pm x/\pm y$ and yaw in $\pm$ directions) with magnitudes $1.0\alpha$\,m and $2.0\alpha^\circ$, respectively.
ICP runs 20 iterations with a 2.0\,m correspondence radius and a 0.7 trimming ratio.
NDT uses a 1.0\,m grid resolution with 20 iterations.
Our method performs class-aware point-to-segment matching with a coarse-to-fine schedule (8 coarse + 15 fine iterations) and class-specific gates (coarse boundary gate 4.0\,m; fine gates 1.5/1.5/1.8\,m for crossing/divider/boundary).
We report per-frame errors averaged over the $K$ perturbations and aggregate over frames using the mean.
\begin{figure*}[t]
  \centering
\includegraphics[width=\linewidth,height=0.3\textheight,keepaspectratio]{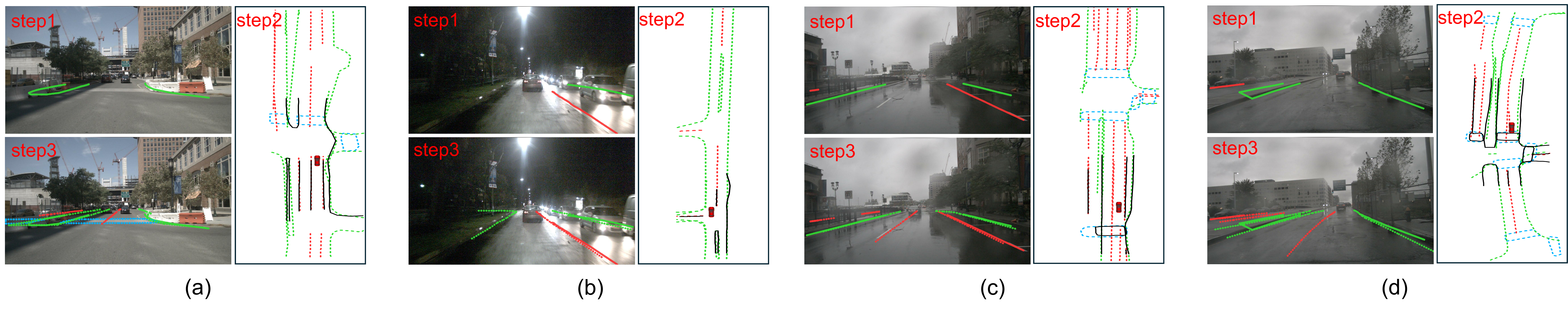}
  \caption{Visualization of LG-FA localization and line completion under diverse conditions on the nuScenes validation split. Each case illustrates how LG-FA aligns the current-frame vector map with the global vector map and recovers missing structures in clear-weather (a), nighttime (b), and rainy scenes (c–d), often improving the continuity of pedestrian crossings, lane dividers, and road boundaries.}
  \label{fig:loc_vis}
  \vspace{0mm}
\end{figure*}
\begin{table}[t]
\centering
\caption{Localization robustness vs. noise scale $\alpha$ ($\sigma_{xy}=1\text{m}\!\times\!\alpha$, $\sigma_{\theta}=2^{\circ}\!\times\!\alpha$). GNSS denotes the noisy initial pose.}
\label{tab:loc_table}
\setlength{\tabcolsep}{3pt} 
\renewcommand{\arraystretch}{1.05} 
\small 
\resizebox{\columnwidth}{!}{%
\begin{tabular}{c|cccc|cccc}
\toprule
\multirow{2}{*}{$\boldsymbol{\alpha}$} &
\multicolumn{4}{c|}{\textbf{Translation error (m)$\downarrow$}} &
\multicolumn{4}{c}{\textbf{Heading error (deg)$\downarrow$}} \\
& GNSS & ICP & NDT & \textbf{Ours} & GNSS & ICP & NDT & \textbf{Ours} \\
\midrule
1 & 2.01 & 1.51 & 0.62 & \textbf{0.33} & 2.57 & 1.22 & 0.96 & \textbf{0.24} \\
2 & 4.02 & 3.27 & 1.29 & \textbf{0.72} & 5.13 & 1.92 & 1.34 & \textbf{0.85} \\
3 & 6.03 & 4.52 & 3.02 & \textbf{1.08} & 7.70 & 2.51 & 2.85 & \textbf{1.59} \\
\bottomrule
\end{tabular}%
}
\end{table}

\subsection{Evaluation of Global Vector Map Quality}
As shown in Table~\ref{tab:map_table}, we report not only the mean but also the median and 90th percentile of map quality metrics to characterize the full error distribution. Lane dividers and road boundaries are stable across most sequences: their Chamfer medians are 0.49\,m and 0.53\,m, and the 90th percentiles remain within 1.28\,m and 1.22\,m, respectively. Pedestrian crossings exhibit a heavier-tailed distribution (median 0.55\,m vs.\ mean 2.16\,m), indicating that a smaller subset of challenging scenes dominates the average; these cases often involve fragmented markings or ambiguous endpoints that lead to larger polyline mismatches. Scale error remains consistently low across all classes: dividers show 1.5/0.9/2.7\% (mean/median/90th), boundaries 1.7/1.0/3.1\%, and pedestrian crossings 3.2/1.1/5.8\%. 
These statistics provide a more comprehensive view of performance and support the use of the constructed global vector map as a reliable geometric prior for subsequent localization and line completion.

Figure~\ref{fig:vis_map} visualizes the same four scenes as in the table. The background is a grayscale point cloud, and vector maps are overlaid in a unified coordinate frame. 
Compared with ground truth, lane dividers and road boundaries show continuous coverage on straights and gentle curves, with curvature changes and road trajectories in agreement. 
At complex intersections, elements such as channelizing islands and crosswalks are also largely aligned, with only minor gaps or endpoint offsets in a few occluded or worn regions. 
For clarity, both ground truth and our results use the same color scheme: \textcolor{red}{red} for lane dividers, \textcolor{green}{green} for road boundaries, and \textcolor{blue}{blue} for pedestrian crossings. 
These visualizations are consistent with the quantitative findings and indicate that the constructed global vector map has reliable geometric consistency and scale stability.

\subsection{Localization Accuracy and Line Completion}
After verifying the geometric consistency of the constructed global vector map, we further evaluate the alignment (pose refinement) accuracy and line completion of LG-FA.
Table~\ref{tab:loc_table} reports localization accuracy under different initialization noise levels, using GNSS as the noisy initialization and comparing LG-FA with ICP~\cite{besl1992method} and NDT~\cite{biber2003normal} as registration-based refinement baselines.
Across all noise levels, LG-FA consistently achieves the lowest translation and heading errors, demonstrating robust convergence even under severe pose perturbations.
While ICP and NDT rapidly degrade as initialization noise increases, LG-FA maintains stable and accurate alignment by leveraging both the global vector map prior and class-constrained matching.
These results indicate that LG-FA achieves reliable alignment under large initialization noise, providing a consistent geometric basis for subsequent line completion.

Fig.~\ref{fig:loc_vis} illustrates four representative cases from the validation split, including one clear-weather, one nighttime, and two rainy scenes.
In all cases, road markings are either missing or barely visible due to adverse weather or illumination.
Each case shows three sequential steps:
\textit{Step1} displays the front-camera view overlaid with the current-frame vector map predicted by the base perception backbone, which appears incomplete due to limited visibility or missing lane markings;
\textit{Step2} shows the BEV view, where the current-frame vector map (black solid lines) is aligned to the global vector map prior (colored dashed lines) and missing regions are completed;
\textit{Step3} projects the enhanced geometry back into the front-camera view, often revealing improved continuity of pedestrian crossings, dividers, and boundaries.
These qualitative examples illustrate how LG-FA uses a lightweight global vector map prior to recover missing geometric context and form a more stable map-aligned representation for foreground perception under degraded visibility conditions.

\subsection{Evaluation of Foreground Perception}
%
\begin{figure}[t]
  \centering
  \includegraphics[width=\linewidth]{figure/fig6_v2.pdf}
  \caption{Completion rate comparison between the pose-only baseline (w/o LG-FA) and the full LG-FA pipeline on the nuScenes validation split (150 scenes). LG-FA consistently improves geometric completeness across all three classes, with substantial gains in pedestrian crossings (+69.0\%), lane dividers (+79.9\%), and road boundaries (+61.1\%).}
  \label{fig:ablation_map}
  \vspace{-1mm}
\end{figure}

To quantify the impact of LG-FA on foreground perception without modifying the base perception backbone, we compare the full LG-FA pipeline against a pose-only mapping baseline. The baseline builds the global vector map by directly transforming and merging per-frame predictions using the raw nuScenes ego poses, i.e., without alignment (map-based pose refinement) and without line completion. We evaluate the completion rate, defined as the ratio of detected geometric elements relative to the maximum achievable coverage, averaged across all validation scenes.

As shown in Fig.~\ref{fig:ablation_map}, LG-FA substantially improves the geometric completeness of foreground perception across all three semantic classes. For pedestrian crossings, the completion rate increases from 30.5\% to 99.5\% (+69.0\%), indicating that the pose-only baseline frequently misses or fragments crossing regions that LG-FA successfully recovers through line completion. Lane dividers show the largest relative improvement, rising from 20.1\% to 100.0\% (+79.9\%), demonstrating that alignment refinement and completion are particularly effective for continuous lane-line structures. Road boundaries improve from 38.9\% to 100.0\% (+61.1\%), reflecting enhanced detection of road edges that are often occluded or weakly visible in single-frame observations.

In summary, LG-FA achieves a +70.0\% overall improvement across all classes. These results demonstrate that the combination of class-constrained localization and line completion enables LG-FA to recover substantially more complete geometric structures than pose-only aggregation, thereby improving the geometric reliability and completeness of foreground perception.

\subsection{LG-FA for Autonomous Driving}
%
\begin{figure}[t]
    \centering
\includegraphics[width=\linewidth,height=0.14\textheight,keepaspectratio]{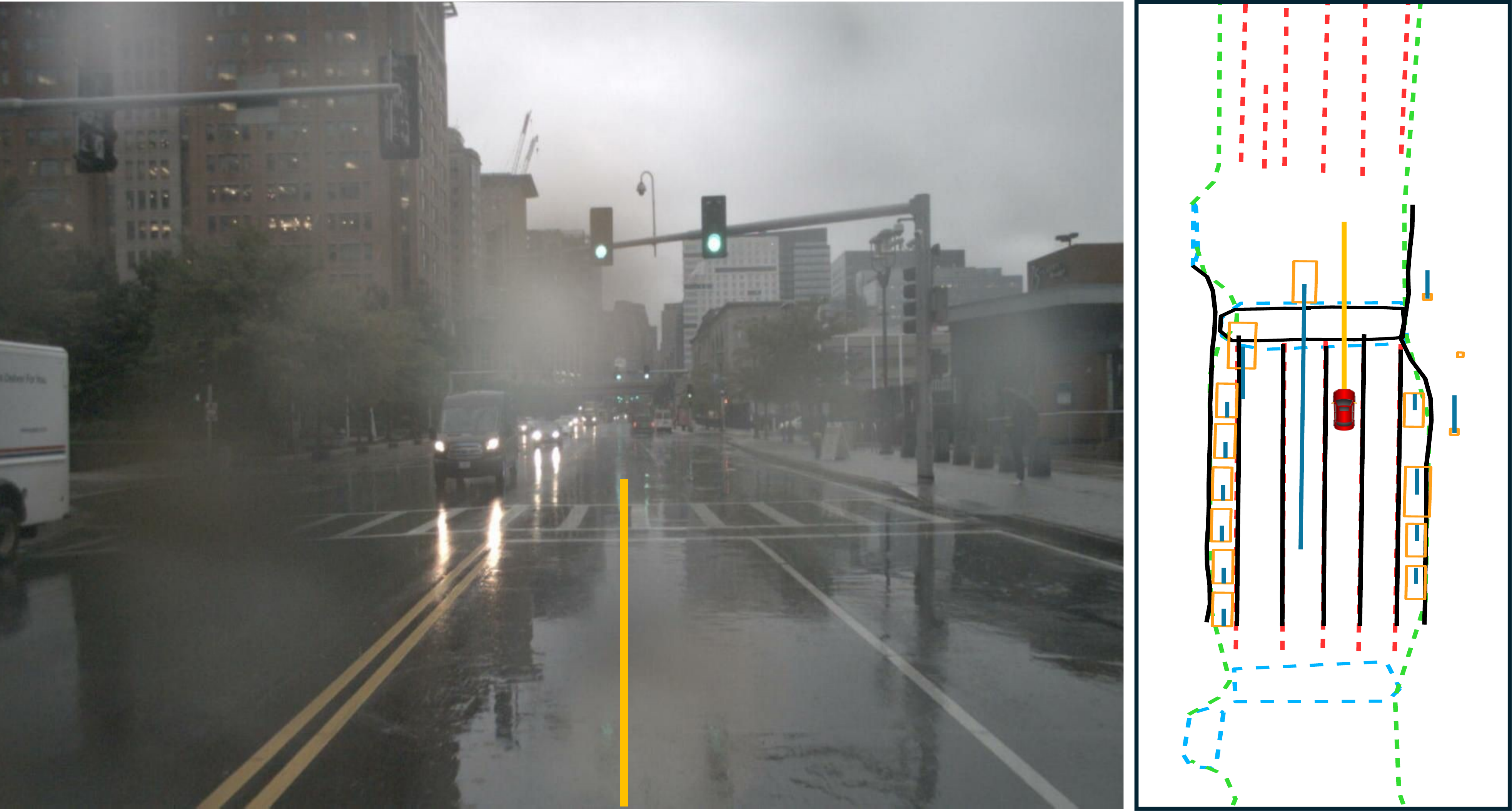}
    \caption{A qualitative example of downstream planning on LG-FA augmented foreground perception in the global map frame. Predicted object trajectories and the planned ego path are overlaid on augmented lanes and road boundaries. Left: front camera view. Right: BEV view.}
    \label{fig:proj_vis}
    \vspace{-1mm}
\end{figure}

In this work, we focus our quantitative evaluation on the geometric completeness, temporal stability, and localization accuracy of the reconstructed foreground, as these properties are critical prerequisites for many downstream autonomy modules.
Leveraging the constructed global vector map and accurate ego localization, LG-FA enables downstream planning to operate in augmented foreground perception. Specifically, after projecting the ego vehicle and detected objects into a unified global coordinate frame, we provide this augmented foreground perception to downstream modules for motion forecasting and ego planning. The predicted object trajectories and planned ego path are then visualized over the augmented global lanes and road boundaries, as illustrated in Fig.~\ref{fig:proj_vis}. The left subfigure presents the front camera view, while the right subfigure shows the corresponding BEV representation.
This visualization highlights the planning behavior expressed in the global map frame, where the planner can leverage a continuous and geometrically consistent structural context—even when lane markings or boundary cues are weak or partially missing in raw sensor observations.
A comprehensive trajectory-level evaluation of downstream decision-making and control performance will be conducted in future work to further validate the system-level benefits of LG-FA.

\section{Conclusion}
Unlike approaches that rely on navigation systems or high-definition maps to complete the current scene, LG-FA is an inference-time module for scene completion in BEV-based end-to-end autonomous driving. It constructs a lightweight background map from historical BEV vectors, refines ego pose via class-constrained alignment, and augments fragmented topology. On nuScenes, LG-FA improves geometric completeness, temporal stability, and localization accuracy, providing stable map-aligned context under incomplete online perception.


{\small
\FloatBarrier
\bibliographystyle{ieeenat_fullname}
\bibliography{references}
}

\end{document}